\documentclass[remotesensing,article,accept,moreauthors,pdftex]{Definitions/mdpi}
\usepackage{xcolor}
\usepackage{amsmath,amssymb,amsfonts}
\usepackage{algorithmic}
\usepackage{graphicx}
\usepackage{textcomp}

\usepackage{amsmath}
\usepackage[lined,boxed]{algorithm2e}
\usepackage{caption}

\usepackage{changes}

\graphicspath{{./Graphs/}}

\definechangesauthor[name={Li Shen}, color=red]{L.}  %

\firstpage{1}
\pubvolume{1}
\issuenum{1}
\articlenumber{0}
\pubyear{2021}
\copyrightyear{2021}
\externaleditor{Academic Editor: {Mohammad Awrangjeb}} 
\datereceived{19 November 2021}
\dateaccepted{13 December 2021}
\datepublished{15 December 2021}
\hreflink{https://doi.org/} 

\Title{S2Looking: A Satellite Side-Looking Dataset for Building Change Detection}

\TitleCitation{S2Looking: A Satellite Side-Looking Dataset for Building Change Detection}

\Author{Li Shen $^{1}$\orcidA{}, Yao Lu $^{1,}$*, Hao Chen $^{2}$\orcidB{}, Hao Wei $^{3}$, Donghai Xie $^{4}$, Jiabao Yue $^{4}$, Rui Chen $^{3}$\orcidC{}, Shouye Lv $^{1}$ \linebreak~and Bitao Jiang $^{1}$}

\AuthorNames{Li Shen, Yao Lu, Hao Chen, Hao Wei, Donghai Xie, Jiabao Yue, Rui Chen, Shouye Lv and Bitao Jiang}

\AuthorCitation{Shen, L.; Lu, Y.; Chen, H.; Wei, H.; Xie, D.; Yue, J.; Chen, R.; Lv, S.; Jiang, B.}

\address{%
$^{1}$ \quad Beijing Institute of Remote Sensing, Beijing 100011, China; shenli@bjirs.org.cn (L.S.); lvshouye@bjirs.org.cn~(S.L.); jiangbitao@bjirs.org.cn (B.J.)\\
$^{2}$ \quad Image Processing Center, School of Astronautics, Beihang University, Beijing 100191, China; justchenhao@buaa.edu.cn\\
$^{3}$ \quad School of Microelectronics, Tianjin University, Tianjin  300072, China;  weihao@tju.edu.cn (H.W.); ruichen@tju.edu.cn (R.C.)\\
$^{4}$ \quad Institute of Resource and Environment, Capital Normal University,  Beijing 100048, China; xiedonghai@cnu.edu.cn (D.X.); yuejiabao2019@cnu.edu.cn (J.Y.)}

\corres{Correspondence: yaolu@bjirs.org.cn}

\abstract{Building-change detection underpins many important applications, especially in
the military and crisis-management domains. Recent methods used for change detection
have shifted towards deep learning, which depends on the quality of its training data. The
assembly of large-scale annotated satellite imagery datasets is therefore essential for global
building-change surveillance. Existing datasets almost exclusively offer near-nadir viewing
angles. This limits the range of changes that can be detected. By offering larger observation
ranges, the scroll imaging mode of optical satellites presents an opportunity to overcome this
restriction. This paper therefore introduces S2Looking, a building-change-detection dataset
that contains large-scale side-looking satellite images captured at various off-nadir angles.
The dataset consists of 5000 bitemporal image pairs of rural areas and more than 65,920 annotated instances of changes throughout the world.
The dataset can be used to train deep-learning-based change-detection algorithms. It
expands upon existing datasets by providing (1) larger viewing angles; (2) large illumination
variances; and (3) the added complexity of rural images. To facilitate {the} use of the dataset,
a benchmark task has been established, and preliminary tests suggest that deep-learning
algorithms find the dataset significantly more challenging  than the closest-competing
near-nadir dataset, LEVIR-CD+. S2Looking may therefore promote important advances in
existing building-change-detection algorithms.}

\keyword{change detection;  remote sensing;  benchmark dataset;  neural networks}

\begin{document}


\section{Introduction}\label{sec:introduction}

Building-change detection underpins a range of applications in domains such as urban
expansion monitoring~\cite{Shunping2019Fully}, land use and cover type change
monitoring~\cite{2011Evaluation,Demir2013Updating}, and resource management and
evaluation~\cite{S2018Land}. It is of particular importance in the contexts of military
surveillance and crisis management~\cite{2010Earthquake}, where changes in buildings may
be indicative of a developing threat or areas in which to focus disaster relief. Change
detection involves identifying changes and differences in an object or phenomenon at
different times~\cite{1989Review}. Remote-sensing-based change detection uses
multitemporal remote-sensing image data to analyze the same area in order to identify
changes in the state information of ground features according to changes in the images.

Many change-detection methods have been proposed over the years. Traditional methods
tended to be either pixel-based~\cite{2016Automatic,David2012Understanding} or
object-based~\cite{2017Object,2017Unsupervised, Masroor2013Change}.
Pixel-based change-detection methods involve pixel by pixel analysis of spectral or textural
information of input image pairs followed by threshold-based segmentation to obtain the
detection results~\cite{2016Automatic,David2012Understanding}.
Object-based change-detection methods similarly rely on spectral and textural information,
but also consider other cues, such as structural and geometric
details~\cite{2017Object,2017Unsupervised}.
However, while these methods can effectively extract geometric structural details and set
thresholds, they are easily influenced by variations in image detail and quality. This
undermines their accuracy~\cite{2015A}. In recent years, the principal methods used for
remote-sensing-based change detection have therefore shifted towards deep learning. This
reflects a wider revolution in computer vision research~\cite{Peng2019End}.
Change-detection methods based on deep-learning include dual-attention fully
convolutional Siamese networks (DASNet)~\cite{2020DASNet}, image fusion networks
(IFN)~\cite{2020Adeeply}, end-to-end change detection based on UNet++
(CD-UNet++)~\cite{Peng2019End}, fully convolutional Siamese networks based on
concatenation and difference (FC-Siam-Conc and FC-Siam-Diff)~\cite{8518015}, and
dual-task constrained deep Siamese convolutional networks
(DTCDSCN)~\cite{2020Building}. Each of these reduces the risk of error by eliminating the
need for preprocessing of the images~\cite{2020Change}.

Although deep-learning-based change-detection methods generally outperform other
change-detection methods, their performance is heavily dependent on the scale, quality, and
completeness of the datasets they use for training. A strong demand has therefore emerged
for large-scale and high-quality change-detection datasets. A number of open datasets for
remote-sensing change detection have been developed to meet this demand, such as the
Change Detection Dataset~\cite{2020AFeature}, the WHU Building
Dataset~\cite{Shunping2019Fully}, the SZTAKI Air Change Benchmark Set
(SZTAKI)~\cite{2009Change,2009AMixed}, the OSCD dataset (OSCD)~\cite{8518015}, the
Aerial Imagery Change Detection dataset (AICD)~\cite{2011Constrained}, and the
LEVIR-CD dataset~\cite{2020ASpatial}, which was released last year.
{
In addition, the privacy concerns can be addressed by privacy-preserving
deep-learning-based techniques
while providing the real-world data from satellite images~\cite{rs13112221}.}

However, most of these change-detection datasets are based on near-nadir imagery. While
this is sufficient for certain kinds of change detection, the observed building features in these
datasets are relatively simple, limiting the scope for change-detection algorithms to serve a
comprehensive range of practical applications.

\begin{figure}[H]
\includegraphics[width=0.6\linewidth]{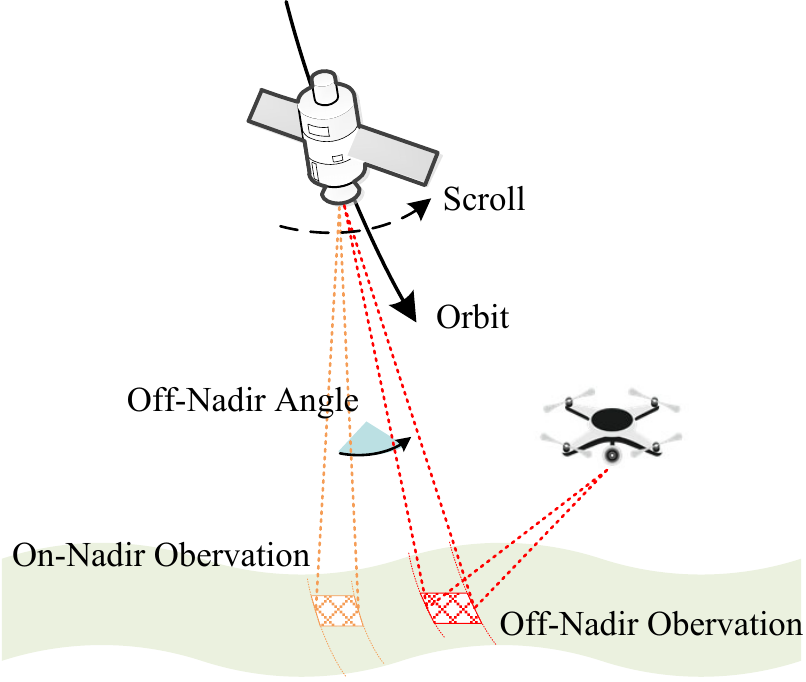}
   \caption{Scroll imaging. The satellite scrolls during the imaging process and obtains
   side-looking remote-sensing images.}
\label{fig:Roll_Imaging}
\end{figure}

A potential solution to this constraint is offered by the scroll imaging mode  {adopted}  by the cameras in a number of modern optical satellites. Figure~\ref{fig:Roll_Imaging} shows the basic way in which this works.
In outline, as an optical satellite pursues its orbit, the onboard high-resolution cameras are
able to scroll so that they can capture multiple images of the same object from different
angles, rather than solely from overhead~\cite{Deren2021}. Unlike near-nadir satellite
imagery, side-looking imagery can capture more relevant details of ground objects and yield
more potentially useful information. Scroll imaging endows surveillance satellites with a
better imaging range and a shorter revisit period than conventional mapping
satellites~\cite{2007Orthophoto}.  For example, the GF-1 satellite has a revisit period of 41
days, together with 4 days for $\pm35^{\circ}$ off-nadir angles; the GF-2 satellite has a
revisit period of 69 days, together with 5 days for $\pm35^{\circ}$ off-nadir angles.

To date, the only study that has made a serious effort to develop a dataset consisting of
multiangle satellite-derived images of specific objects in this way is SpaceNet
MVOI~\cite{2020SpaceNet}. However, this dataset is not focused on change detection, so it
does not offer bitemporal images, and only contains 27 views from different angles at a
single geographic location.

\begin{figure}[H]
\includegraphics[width=0.9\linewidth]{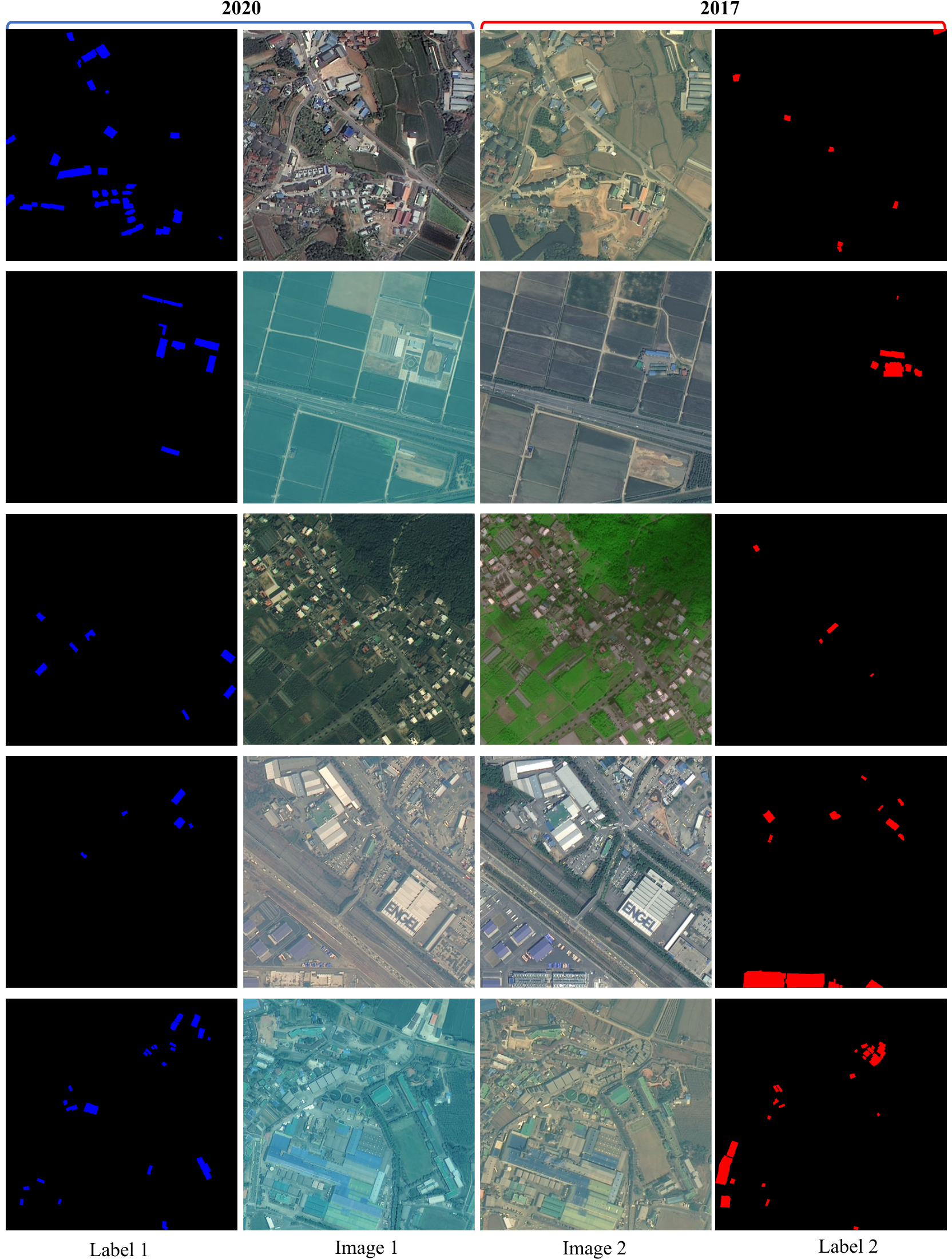}
   \caption{Samples from the S2Looking dataset. Images 1 and 2 in
   Figure~\ref{fig:annotation} are bitemporal remote-sensing images, while Labels 1 and 2 are
   the corresponding annotation maps. Labels 1 and 2 indicate pixel-precise newly built and
   demolished areas of buildings, respectively.}
\label{fig:annotation}
\end{figure}

In this paper, we therefore introduce S2Looking, a building-change-detection dataset that
contains large-scale side-looking satellite images captured at various off-nadir angles. The
dataset consists of 5000 bitemporal pairs of very-high-resolution (VHR) registered images
collected from the GaoFen (GF), SuperView (SV) and BeiJing-2 (BJ-2) satellites from 2017 to
2020. The imaged areas cover a wide variety of globally-distributed rural areas, with very
different characteristics, as can be seen in Figure~\ref{fig:annotation}.
{The dataset shown in \mbox{Figure~\ref{fig:annotation}} contains various scenes from all
over the world, including villages, farms, villas, retail centers, and industrial areas, which
relate to each row above, respectively.
}
Each image in the pairs is $1024\times1024$ with an image resolution of 0.5$\sim$0.8 m/pixels.
{
The image pairs in the dataset are converted from the original TIFF format with 16 bit to
PNG format with 8~bit.
}
The pairs are accompanied by 65,920 expert-based annotations of changes and two label maps that separately indicate newly built and demolished building regions for each sample in the dataset.
The side-view imaging and complex rural scenes in the dataset present whole new
challenges for change-detection algorithms by making the identification and matching of
buildings notably more difficult.
Placing higher requirements on an algorithm's robustness by confronting it with more
complex ground targets and imaging conditions increases its practical value if it can
successfully meet such challenges. Thus, S2Looking offers a whole new resource for the
training of deep-learning-based change-detection algorithms.
It significantly expands upon the degree of richness offered by available datasets, by
providing (1) larger viewing angles; (2) large illumination variances; and (3) the added
complexity of the characteristics encountered in rural areas. It should also be noted that
algorithms trained on S2Looking are likely to be better able to deal with other forms of
aerially-acquired imagery, for instance by aircraft, because such images present similarly
offset viewing angles. To facilitate use of the dataset, we have established a benchmark task
involving the pixel-level identification of building changes in bitemporal images. We have
also conducted preliminary tests of how existing deep-learning algorithms might perform
when using the dataset. When this was compared with their performance on the
closest-competing dataset, LEVIR-CD+, which is based on near-nadir images, the results
revealed that S2Looking was substantially more challenging.
This suggests it has the potential to induce step-change developments in
building-change-detection algorithms that seek to address the challenges it presents.

In the next two subsections, we discuss work relating to change detection based on
remote-sensing images and change-detection datasets, with the latter playing an important
role in the analysis and processing required for the prior.

\subsection{Change Detection Methods}


Traditional change-detection methods were originally either
pixel-based~\cite{David2012Understanding,2016Automatic} or
object-based~\cite{2017Unsupervised,Masroor2013Change,2017Object}.
Traditional methods of change detection in remote-sensing images are designed on the basis
of handcrafted features and supervised classification algorithms.
These methods were capable of extracting geometric structural features from images and then applying thresholds that would indicate some kind of change when they were cross-compared. However, as pointed out in \cite{2015A}, these kinds of methods are very susceptible to variations in the images and their quality, making their accuracy questionable.
However, with the rapid development of computer hardware and artificial intelligence, deep-learning-based methods have also been widely studied and applied.
When methods based on deep learning began to hold sway in image processing, it became apparent that they could be used to address these issues, leading to their rapid adoption in change detection.

Deep-learning-based change detection can be roughly divided into metric-based methods
and classification-based methods. Metric-based methods involve learning a parameterized
embedding space where there is a large distance between the changed pixels and a small
distance between the unchanged pixels. A change map can then be obtained by calculating
the distances between embedded vectors at different times in the same position. Zhan et
al.~\cite{8022932} used a deep Siamese fully convolutional network with weight sharing to
learn the embedding space and extract features from images captured independently at
different times. Saha et al.~\cite{2019Unsupervised} proposed an unsupervised deep change
vector analysis method based on a pretrained convolutional neural network (CNN) and
contrastive/triplet loss functions~\cite{2020Adeep,Zhang2019Triplet}. Chen et
al.~\cite{2020DASNet} proposed DASNet to overcome the influence of pseudo change
information in the recognition process. Chen et al.~\cite{2020ASpatial} proposed a
spatiotemporal attention neural network (STANet) based on the FCN-network
(STANet-Base). This enabled them to produce two improved models with self-attention
modules; one with a basic spatiotemporal attention module (STANet-BAM), the other with a
pyramid spatiotemporal attention module (STANet-PAM).

Classification-based methods typically use CNNs to learn the mapping from bitemporal data
to develop a change probability map that classifies changed categories in a pixelwise fashion.
A changed position has a higher score than an unchanged position. Zhang et
al.~\cite{2020Adeeply} proposed the IFN model, which relies on a deep supervised
difference discrimination network (DDN) to detect differences in the proposed image
features. \mbox{Peng et al.~\cite{Peng2019End}} developed an improved automatic coding
structure based on the UNet++ architecture and proposed an end-to-end detection method
with a multilateral fusion strategy. Daudt et al.~\cite{8451652,8518015} proposed a fully
convolutional early fusion (FC-EF) model, which concatenates image pairs before passing
them through a UNet-like network. They also developed the FC-Siam-Conc and
FC-Siam-Diff models, which concatenate image pairs after passing them through a Siamese
network structure. Liu et al.~\cite{2020Building} proposed a dual-task constrained deep
Siamese convolutional network DTCDSCN, within which spatial and channel attention
mechanisms are able to obtain more discriminatory features. This network also incorporates
semantic segmentation subnetworks for multitask learning. Chen et al.~\cite{Chen2021}
proposed a simple yet effective change-detection network (CDNet) that uses a deep Siamese
fully convolutional network as the feature extractor and a shallow fully convolutional
network as the change classifier for feature differences in the images. Very recently, Chen et
al.~\cite{chen2021efficient} proposed an efficient transformer-based model (BiT), which
leverages a transformer to exploit the global context within a token-based space, thereby
enhancing the image features in the pixel space.

The main difference between metric-based and classification-based approaches is the network structure.
A classification-based approach offers more diverse choices than a metric-based one. For
example, the former mostly makes use of a late fusion architecture, where the bitemporal
features are first extracted and then compared to obtain the change category. The latter,
however, can make use of both late and early fusion. As a result, classification-based
methods are more commonly used than metric-based methods. In terms of the loss function,
the former usually uses contrastive loss, which is designed for paired bitemporal feature
data, while the latter uses cross-entropy loss for the fused features. Overall, deep learning
approaches are very robust when it comes to variability in the data, but one of their key
drawbacks is that they are only as good as the datasets that they use for
training~\cite{2019DeepM, 2019DeepL, Jiang2020PGA}. This places a heavy emphasis upon
the development of high-quality and comprehensive datasets, with change detection being
no exception.
%
\subsection{{Change Detection Datasets}}

There are many large-scale benchmark datasets available for detection, recognition, and
segmentation based on everyday images. These include ImageNet~\cite{2009ImageNet},
COCO~\cite{2014Microsoft}, PIE~\cite{2020PIE}, MVTec AD~\cite{2020MVTec},
Objects365\cite{2020Objects365}, WildDeepfake~\cite{2020WildDeepfake},
FineGym~\cite{2020FineGym}, and ISIA Food-500~\cite{min2020isia}. There are also
large-scale datasets containing satellite and aerial imagery, such as ReID~\cite{2019Vehicle},
which contains 13 k ground vehicles captured by unmanned aerial vehicle (UAV) cameras,
and MOR-UAV~\cite{moruav2020}, which consists of 30 UAV videos designed to localize
and classify moving ground objects. WHU~\cite{LiuJ20} consists of thousands of multiview
aerial images for multiview stereo (MVS) matching tasks. DeepGlobe~\cite{8575485}
provides three satellite imagery datasets, one for building detection, one for road extraction,
and one for land-cover classification. xBD~\cite{gupta2019xbd} and
Post-Hurricane~\cite{chen2018benchmark} consist of postdisaster remote-sensing imagery
for building damage assessment. SensatUrban~\cite{hu2021semantic} contains urban-scale
labeled 3D point clouds of three UK cities for fine-grained semantic understanding.
FAIR1M~\cite{sun2021fair1m} provides more than 1 million annotated instances labeled
with their membership in 5 categories and 37 subcategories, which makes it the largest
dataset for object detection in remote-sensing images presented so far.
{
The stereo satellite imagery~\cite{ijgi10100697} and light detection and ranging (LiDAR)
point clouds~\cite{rs13234766} provide the base data for the 3D city modeling technology.}
However, in the above datasets, each area is covered by only a single satellite or aerial image.
Change-detection models need to be trained on datasets consisting of bitemporal image
pairs, which typically correspond to different sun/satellite angles and different atmospheric
conditions.

In Table~\ref{tab:data-table}, we present the statistics relating to existing change-detection
datasets, together with those for our S2Looking dataset. It can be seen that
SZ-TAKI~\cite{2009Change,2009AMixed} contains 12 optical aerial image pairs. These focus
on concerns such as building changes and the planting of forests. OSCD~\cite{8518015}
focuses on urban regions and includes 24 multispectral satellite image pairs. Unlike other
datasets, AICD~\cite{2011Constrained} is a synthetic change-detection dataset containing
500 image pairs. The LEVIR-CD dataset~\cite{2020ASpatial} consists of 637
manually-collected image patch pairs from Google Earth. Its recently expanded version,
LEVIR-CD+, contains 985 pairs. The Change Detection Dataset~\cite{2020AFeature} is
composed of 11 satellite image pairs. The WHU Building Dataset~\cite{Shunping2019Fully}
consists of just one max-width aerial image collected from a region that suffered an
earthquake and was then rebuilt in the following years. Our own dataset, S2Looking,
contains 5000 bitemporal pairs of rural images. Generally, S2Looking has the most image
pairs; WHU has the largest size and best resolution; S2Looking offers the most change
instances and change pixels. Apart from these general change-detection datasets, a river
change-detection dataset~\cite{2019GETNET} has also been released that specifically
concentrates on the detection of changes in rivers that are locatable through hyperspectral
images.

\end{paracol}
\nointerlineskip
\begin{specialtable}[H]
\setlength{\tabcolsep}{3.0mm}
\widetable
\caption{Statistical characteristics of existing change-detection datasets.}
\label{tab:data-table}
\scalebox{.8}[0.8]{\begin{tabular}{ccccccc}
\midrule
\textbf{Dataset}  & \textbf{Pairs} & \textbf{Size}      & \textbf{Is Real?} & \textbf{Resolution}\textbf{/m} & \textbf{Change Instances} & \textbf{Change Pixels} \\ \midrule
SZTAKI~\cite{2009Change,2009AMixed}   & 12    & 952~$\times$~640   & $\surd$  & 1.5          & 382              & 412,252       \\
OSCD~\cite{8518015}     & 24    & 600~$\times$~600   & $\surd$  & 10           & 1048            &
148,069       \\
AICD~\cite{2011Constrained}     & 500   & 600~$\times$~800   & $\times$ & none         & 500              & 203,355       \\
LEVIR-CD~\cite{2020ASpatial} & 637   & 1024~$\times$~1024 & $\surd$  & 0.5          & 31,333           & 30,913,975    \\
\multirow{2}{*}{Change Detection Dataset~\cite{2020AFeature}} &
  \multirow{2}{*}{7/4} &
  \multirow{2}{*}{\begin{tabular}[c]{@{}c@{}}4725~$\times$~2700/\linebreak1900~$\times$~1000\end{tabular}} &
  \multirow{2}{*}{$\surd$} &
  \multirow{2}{*}{0.03 to 1} &
  \multirow{2}{*}{\begin{tabular}[c]{@{}c@{}}1987/\linebreak 145\end{tabular}} &
  \multirow{2}{*}{\begin{tabular}[c]{@{}c@{}}9,198,562/\linebreak400,279\end{tabular}} \\
         &       &           &          &              &                  &               \\
\multirow{2}{*}{WHU Building Dataset~\cite{Shunping2019Fully}} &
  \multirow{2}{*}{1} &
  \multirow{2}{*}{32,507~$\times$~15,354} &
  \multirow{2}{*}{$\surd$} &
  \multirow{2}{*}{0.075} &
  \multirow{2}{*}{2297} &
  \multirow{2}{*}{21,352,815} \\
         &       &           &          &              &                  &               \\ \midrule
LEVIR-CD+ & 985   & 1024~$\times$~1024 & $\surd$  & 0.5          & 48,455           & 47,802,614    \\
S2Looking   & 5000  & 1024~$\times$~1024 & $\surd$        & 0.5$\sim$0.8 & 65,920           & 69,611,520 \\
\midrule
\end{tabular}}
\end{specialtable}

\begin{paracol}{2}
\switchcolumn

The datasets described above (excepting our new S2Looking) have played an important part
in promoting the development of change-detection methods. However, they share a common
drawback in that most of the images they contain have been captured at near-nadir viewing
angles. The absence of large-scale off-nadir satellite image datasets limits the scope for
change-detection methods to advance to a point where they can handle more subtle
incremental changes in objects such as buildings. This is the underlying motivation behind
the development of the S2Looking dataset, as described in more detail below.

\subsection{{Contributions}}

Overall, the primary contributions of this paper are as follows: (1) a pipeline for constructing
satellite remote-sensing image-based building-change-detection datasets; (2) presentation of
a unique, large-scale, side-looking, remote-sensing dataset for building-change detection
covering rural areas around the world; (3) a benchmark test upon which existing algorithms
can assess their capacity to undertake the monitoring of building changes when working
with large-scale side-looking images; (4) a preliminary evaluation of the added complexity
presented by the dataset. On the basis of this, we reflect briefly upon potential future
developments in building-change detection based on surveillance satellites.


\section{Materials and Methods}\label{sec:dataset}

As noted in~\cite{Jiang2020PGA}, the lack of public large-scale datasets that cover all kinds
of satellites, all kinds of ground surfaces, and from a variety of angles, places limits upon
how much progress can be made in change-detection algorithm research and the resulting
applications. This has driven our development of the S2Looking dataset. The situation was
somewhat improved by the release of the LEVIR-CD dataset in 2020~\cite{2020ASpatial},
which presented a good range of high-resolution change instances.  LEVIR-CD has very
recently been superseded by LEVIR-CD+ (also {available at} 
 \url{https://github.com/S2Looking/}, {Accessed 1 November 2021}), with an even larger number of bitemporal image
 pairs. However, the LEVIR-CD+ dataset mainly targets urban areas as captured in survey or
 mapping satellite data from Google Earth. It also retains the focus of prior datasets upon
 near-nadir imagery. As LEVIR-CD+ was, until the development of S2Looking, the richest
 dataset available for testing change-detection algorithms, S2Looking has, in many ways,
 been framed against it, hence the focus in S2Looking upon mainly rural targets captured by
 surveillance or reconnaissance satellites at varying off-nadir angles. S2Looking thus
 provides both the largest and the most challenging change-detection dataset available in the
 public domain.
Throughout the remainder of the paper, active comparisons are made between S2Looking
and LEVIR-CD+ because the latter constitutes the immediately preceding state-of-the-art.
Over the course of this section, we look in more detail at the objectives we were pursuing in
developing S2Looking and its data-processing pipeline.

\subsection{Motivation for the New S2Looking Dataset}

As noted above, S2Looking was actively developed to expand upon the change-detection
affordances of LEVIR-CD+. LEVIR-CD+ is itself based on the LEVIR-CD dataset presented in
\cite{2020ASpatial}. In comparison with the 637 image patch pairs in the LEVIR-CD dataset,
LEVIR-CD+ contains more than 985 VHR (0.5 m/pixel) bitemporal Google Earth images, with
a size of $1024 \times 1024$ pixels. These bitemporal images are from 20 different regions
located in several cities in the state of Texas in the USA. The capture times of the image data
vary from 2002 to 2020. Images of different regions were taken at different times. Each pair of
bitemporal images has a time span of 5 years.

Unlike the LEVIR-CD+ dataset, S2Looking mostly targets rural areas. These are spread
throughout the world and were imaged at varying large off-nadir angles by optical satellites.
The use of Google Earth by LEVIR-CD+ actually places certain limits upon it, because, while
Google Earth provides free VHR historical images for many locations, its images are
obtained by mapping satellites to ensure high resolution and geometric accuracy. In contrast,
images captured by optical satellites, especially surveillance satellites, can benefit from their
use of scroll imaging to achieve a better imaging range and a shorter revisit period (as shown
in Figure~\ref{fig:Roll_Imaging}). For example, when a disaster occurs, scroll imaging by
satellites is used to support analysts who require quick access to satellite imagery of the
impacted region, instead of waiting until a satellite reaches orbit immediately above. By
varying the camera angle on subsequent passes, it is also possible to revisit the impacted
region more often to monitor recovery and look for signs of further events. For military
applications, scroll imaging by satellites is frequently used to obtain a steady fix on war
zones~\cite{Hern2021Design}. However, a challenge arising from using off-nadir angles that
needs to be properly met by new change-detection algorithms is that there is a lack of
geometric consistency between tall objects in the gathered satellite images. This effect can be
seen in Figure~\ref{fig:side-look}.
{
The subvertical objects highlighted in yellow in Figure~\ref{fig:side-look} become more
visible and have a larger spatial displacement proportionate with the off-nadir angle in the
middle and right images.}

\begin{figure}[H]
\includegraphics[width=0.98\linewidth]{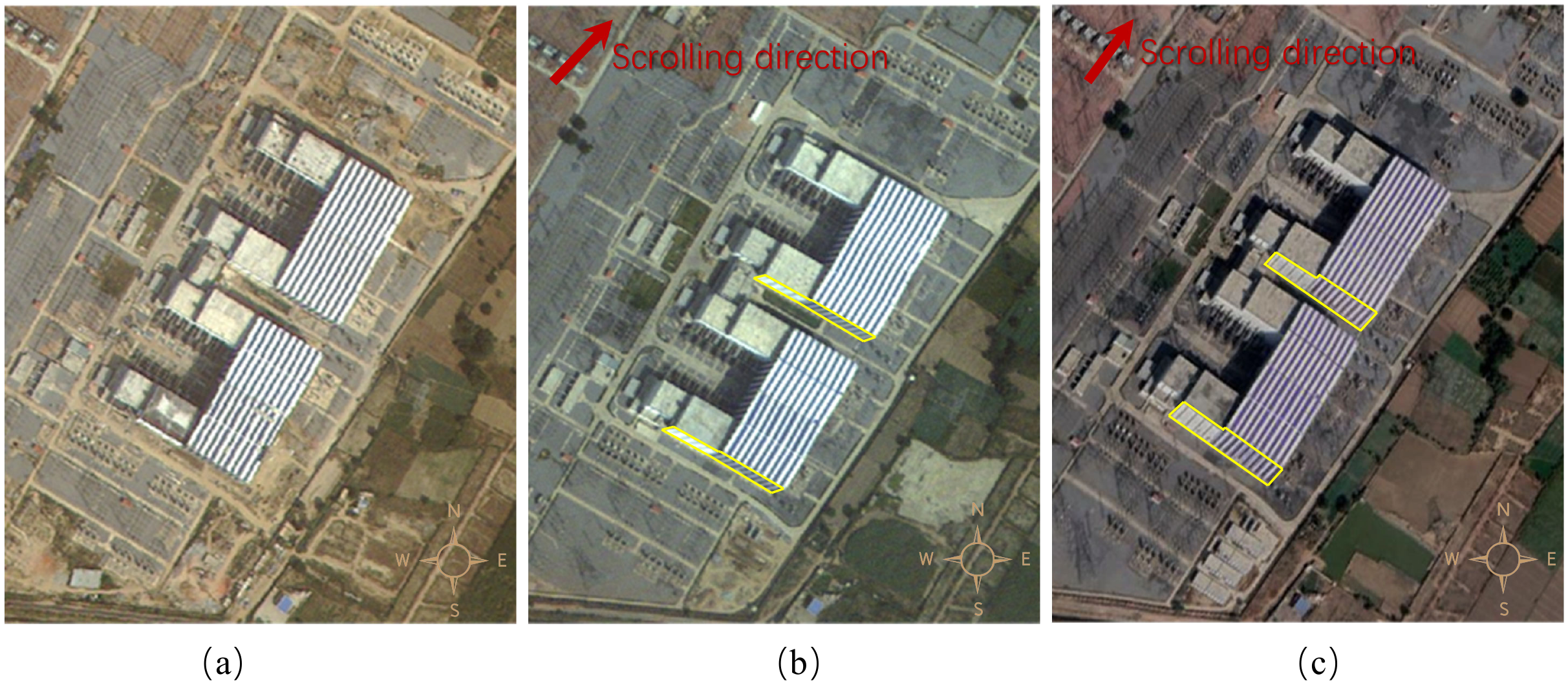}
    \vspace{-4pt}
   \caption{Examples of side-looking remote-sensing images.
     (\textbf{a}) Near-nadir imagery from Google Earth.
     (\textbf{b}) Side-looking imagery at an off-nadir angle of $10^{\circ}$ from S2Looking.
     (\textbf{c}) Side-looking imagery at an off-nadir angle of $18^{\circ}$ from S2Looking.
  }
\label{fig:side-look}
\end{figure}

A further value of developing an off-nadir-based dataset for change detection is that aerial
imagery, airborne and missile-borne, also tends to be captured at large off-nadir angles to
maximize visibility according to the flight altitude (see, for instance, \cite{Kunwar2018}). As
such imagery presents the same problem of there being a lack of geometric consistency
caused by the high off-nadir angles, an image processing model trained on side-looking
satellite imagery will more easily adapt to aerial imagery, raising the possibility of it
supporting joint operations involving satellites and aircraft.

Figure~\ref{fig:location} illustrates the geospatial distribution of our new dataset. Most
remote-sensing images in S2Looking relate to rural areas near the marked cities. The chosen
cities have been a particular focus of satellite imagery, so it was easier to acquire bitemporal
images for these locations. Together, the adjacent rural areas cover most types of rural
regions around the world.

\begin{figure}[H]
\includegraphics[width=0.98\linewidth]{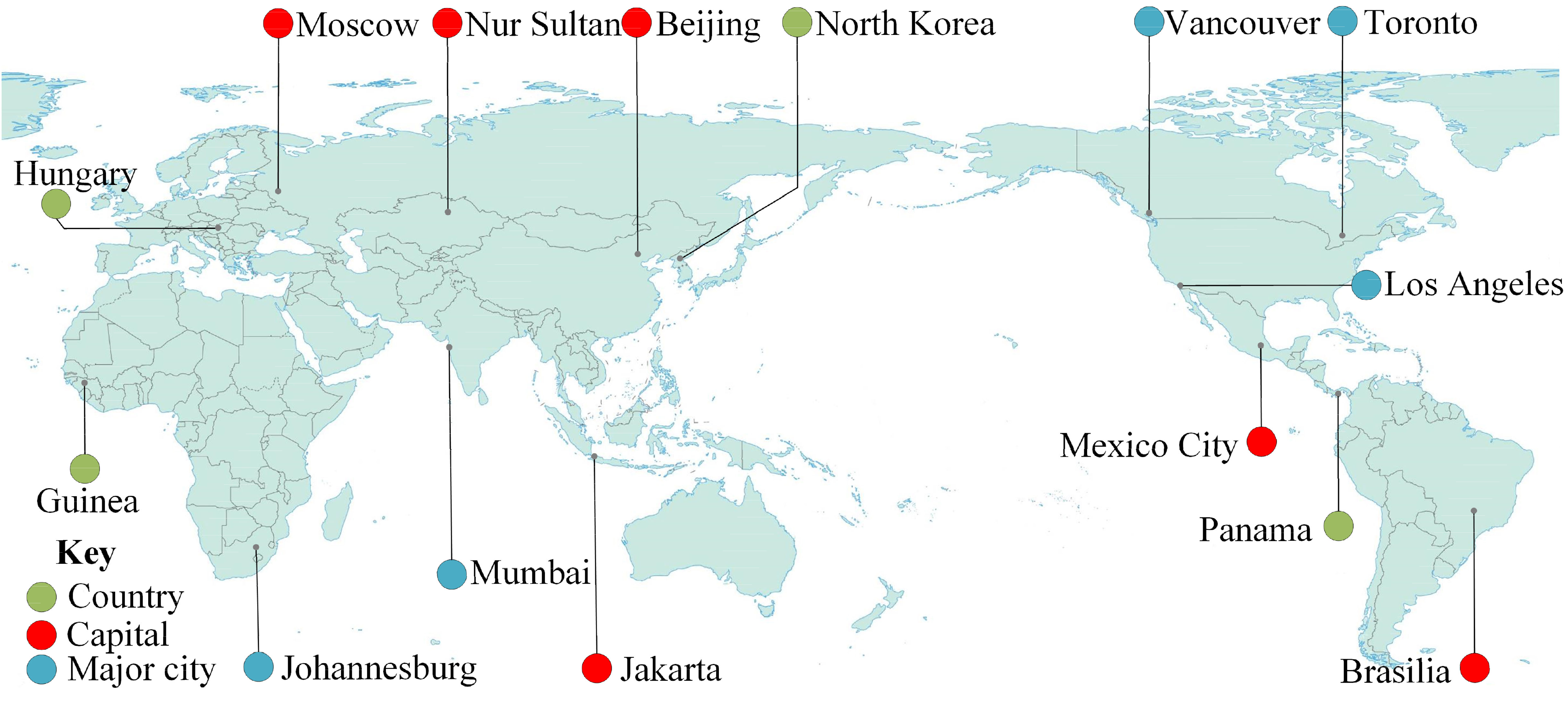}
   \caption{The global locations covered by the bitemporal images in the S2Looking dataset.}
\label{fig:location}
\end{figure}

When it comes to remote sensing change detection, data from rural areas generally have
more value than data from urban areas. There are several reasons for this. First of all, in the
case of military surveillance, sensitive installations are usually built in isolated areas for
safety reasons~\cite{Deren2021}. In the case of disaster rescue, satellite images of remote
rural regions can be obtained faster than aerial photographs because satellites do not have a
restricted flying range~\cite{2020Using}. Therefore, rural images offer the scope to train
remote-sensing change-detection models that can find and update the status of military
installations or destroyed buildings for disaster rescue teams. As is noted in Section 4
regarding the challenges posed by S2Looking, there are certain characteristics that accrue to
rural images that can render it more difficult to recognize buildings, enhancing their value
for training.

{However}, side-looking satellite imagery of rural areas is more difficult to collect than
vertical-looking imagery of urban areas. The buildings can be imaged by satellites from
different sides and projected along different angles into 2D images, as we see in
\mbox{Figure~\ref{fig:side-look}}. Table~\ref{tab:dataset-table} provides a summary of our dataset.
The off-nadir angles
have an average absolute value of $9.86^{\circ}$ and a standard deviation of
$12.197^{\circ}$. A frequency histogram of the dataset is also given in
Figure~\ref{fig:freqroll angle}. Here, it can be seen that the off-nadir angles ranged from
${-35}^{\circ}$ to ${+40}^{\circ}$, with the highest frequency of 1158 clustering between
${-5}^{\circ}$ and ${0}^{\circ}$, reflecting the optimal imaging angle.
Optical satellites usually have on-nadir observation angles of within ${\pm15}^{\circ}$ to maximize their lifespan. The off-nadir mode typically relates to observation angles larger than ${15}^{\circ}$. Our dataset consists of 71.9\% on-nadir and 28.1\% off-nadir image pairs.
This level of variation makes it difficult for a registration algorithm to match feature points between the bitemporal images.
Additionally, irrelevant structures, such as greenhouses and small reservoirs, can interfere
with the building annotation process.

\begin{figure}[H]
\includegraphics[width=0.8\linewidth]{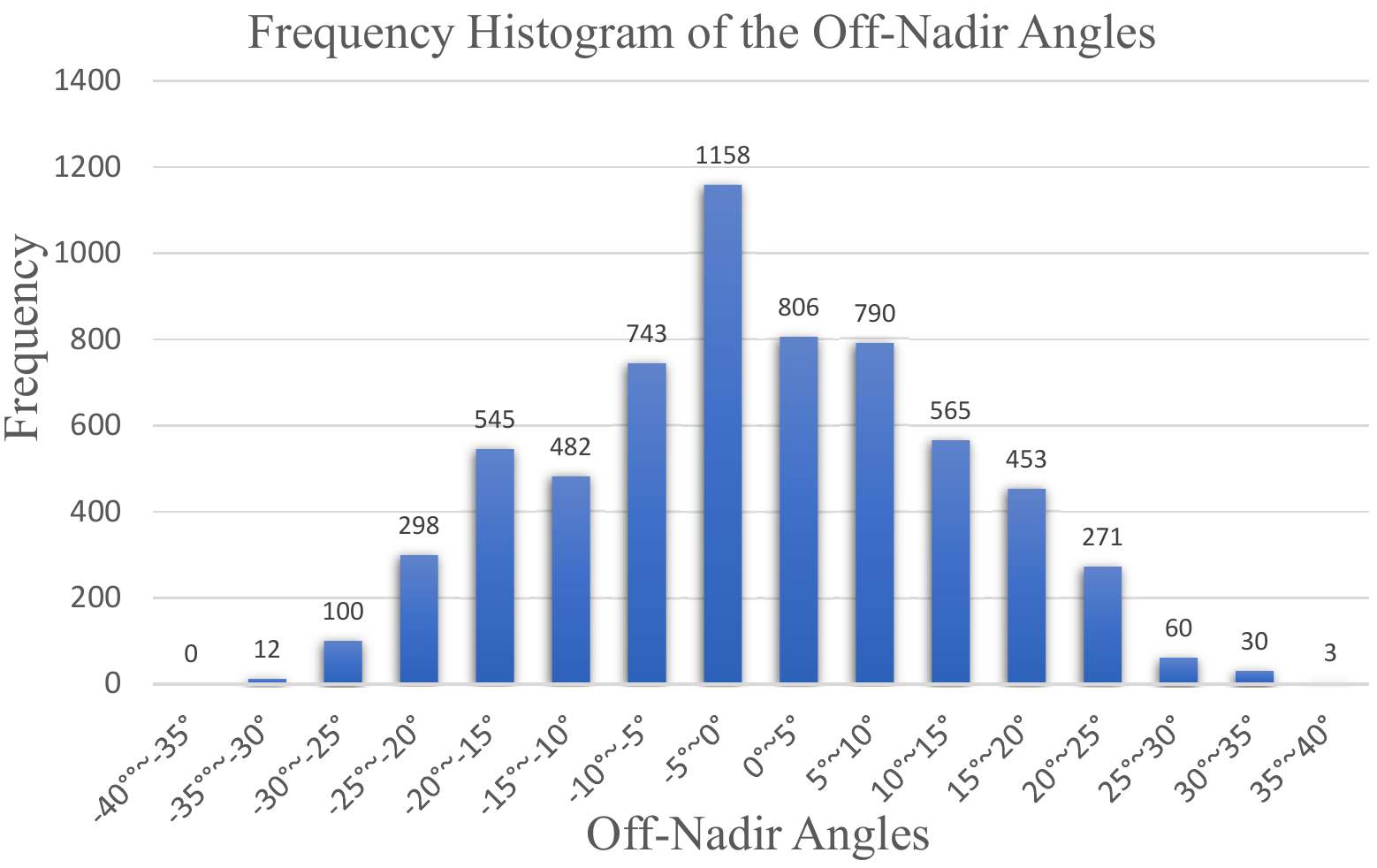}
   \caption{Frequency histogram of the off-nadir angles of the satellite images in the S2Looking dataset.
}
\label{fig:freqroll angle}
\end{figure}

\vspace{-8pt}

\begin{specialtable}[H]
\setlength{\tabcolsep}{10.7mm}
\caption{A summary of the S2Looking dataset.}
\label{tab:dataset-table}
\scalebox{.9}[0.9]{\begin{tabular}{ccc}
\toprule
\textbf{Type}                 & \textbf{Item}             & \textbf{Value}  \\
\midrule
\multirow{5}{*}{Image Info}  & Total Image Pairs      & 5000             \\
                              & Image Size                & 1024~$\times$~1024      \\
                              & Image Resolution          & 0.5$\sim$0.8 m     \\
                              & Time Span                 & 1$\sim$3 years \\
                              & Modality                  & RGB image       \\
                              & {Image Format}                  & {PNG 8bit}       \\
                              \midrule
\multirow{4}{*}{Off-Nadir Angle Info} &Average Absolute Value& 9.861$^{\circ}$ \\
                              & Median Absolute Value & 9.00$^{\circ}$    \\
                              & Max Absolute Value & 35.370$^{\circ}$    \\
                              & Standard Deviation    & 12.197$^{\circ}$  \\
                              \midrule
\multirow{2}{*}{Accuracy Info} &Registration Accuracy& $ \leq8$ pixels \\
                              & Annotation Accuracy & $ \leq2$ pixels  \\
                              \bottomrule

\end{tabular}}
\end{specialtable}

\subsection{The $\text{S2Looking}$ Data Processing Pipeline}\label{sec:pipeline}
An illustration of the data-processing pipeline for our S2Looking dataset is shown in
Figure~\ref{fig:pipeline}. This pipeline is discussed in greater detail below:

\begin{figure}[H]
\includegraphics[width=0.7\linewidth]{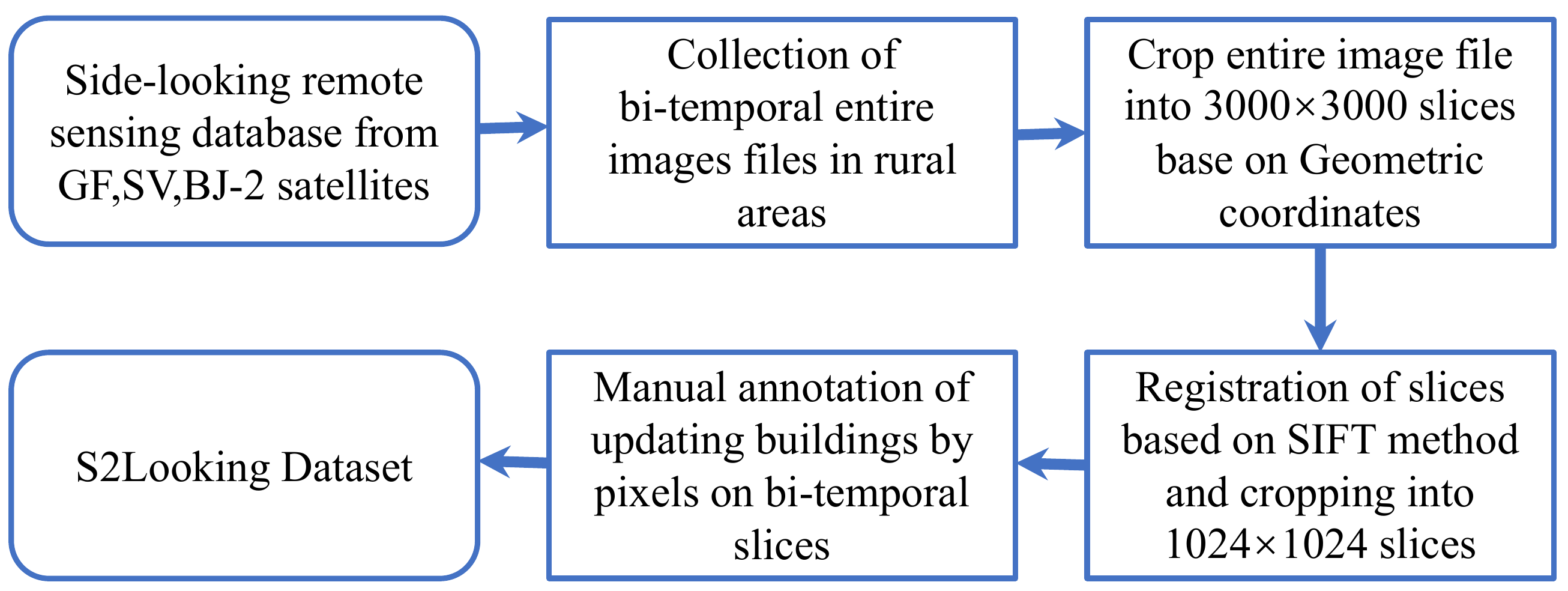}
   \caption{Data processing pipeline for the S2Looking dataset.}
\label{fig:pipeline}
\end{figure}

{\textbf{Image Selection.}}
 The 
 images in the dataset were selected on the basis of both time and location. Locations were
 selected that covered typical types of rural areas around the world. The satellite images were
 then checked, and only locations that had been captured more than once in the past 10 years
 were retained.

{\textbf{Collection of bitemporal image files.}}
Based on the chosen rural locations, we next checked our satellite image database and
collected time-series images for each specific location. We then selected entire bitemporal
image files with an intersection-over-union (IoU) greater than 0.7 and a time span greater
than 1 year.

{\textbf{Cropping into slices based on geometric coordinates.}}
Each remote-sensing image contained the rough coordinate information for each pixel, which
was obtained by geometric inversion of the position of the satellite and the camera angle.
Based on these geometric coordinates, the intersecting areas were selected and cropped into $3000\times3000$ slices.
The final image size of $1024\times1024$ is better suited to GPU acceleration because it
matches the number of GPU threads. The accuracy of the coordinates was about 20 m (about
40$\sim$100 pixels), so we cropped the intersection areas into $3000\times3000$ slices with
a stride of 1024 for the registration. The $3000\times3000$ slices therefore overlapped. After
the registration process, the central $1024\times1024$ area was retained to form the
bitemporal pairs in the S2Looking dataset.

{\textbf{Bitemporal image registration.}}
Image registration is essential for the preparation of image pairs for a change-detection
dataset~\cite{2004Earthquake}. Although accurate spatial coregistration for images captured
at near-nadir angles can be achieved with accurate digital elevation models
(DEMs)~\cite{2018ARelative,2018AOrbit}, the geolocalization accuracy deteriorates for large
off-nadir angles, especially in the case of hilly/mountainous areas~\cite{2012Review,
2015Change}. In addition, dense global DEMs (i.e., with a resolution of at least 2 m) were not
available.

{According to our experience, the building-change-detection task is achievable when the
precision of the geometric alignment of bitemporal image pairs is less than 8 pixels.}
We therefore used a scale-invariant feature transform (SIFT)
algorithm~\cite{2004Distinctive} to find and match the feature points in each pair of
corresponding images. The SIFT algorithm constructs a multiscale image space by means of
Gaussian filtering and searches for extreme points in the difference of Gaussian (DOG)
image, which are used as feature points. For the feature description, the SIFT algorithm uses
a gradient direction histogram to describe the feature points and uses the ratio of the
distances to the nearest and second nearest points as the basis for matching.
The SIFT algorithm is robust to illumination changes and can even handle some degree of affine or perspective distortion~\cite{12346}. To improve the accuracy of the SIFT algorithm, we used floating-point variables to store the gray values of each image. This avoids ending up with missing values because of integer approximation when the Gaussian filtering and DOG results are calculated.

After this, incorrectly matched pairs were deleted according to a random sample consensus (RANSAC) homography transformation~\cite{2018Image}. Finally, based on the matched points, the image pairs were resampled using a homography matrix to ensure the matched points had the smallest possible pixel distances in each image, as shown in Figure~\ref{fig:registration}. Thus, the same features and buildings in the image pairs had the same pixel index in the bitemporal images and were suitable for change detection by recognizing any inconsistent buildings.

To avoid misaligned image pairs and guarantee the registration accuracy, all the bitemporal
images in the S2Looking dataset were manually screened by remote sensing
image-interpretation experts to ensure that the maximum registration deviation was less
than 8 pixels. This means that misaligned areas are less than 1/16 of the average building
change size in S2Looking (1056 pixels). Although absolutely accurate registration is probably
impossible, further improvement of the registration of S2Looking is required. This would
enhance the performance of change-detection models trained on the dataset. Registration
could be improved, for instance, by manual annotation of the same points in the bitemporal
pairs.
A large number of registration methods are now available based on deep learning, and our
hope is that the process can be iteratively refined by various researchers making use of the
S2Looking dataset and undertaking their own registration activities prior to applying
change-detection methods.
\begin{figure}[H]
\includegraphics[width=0.8\linewidth]{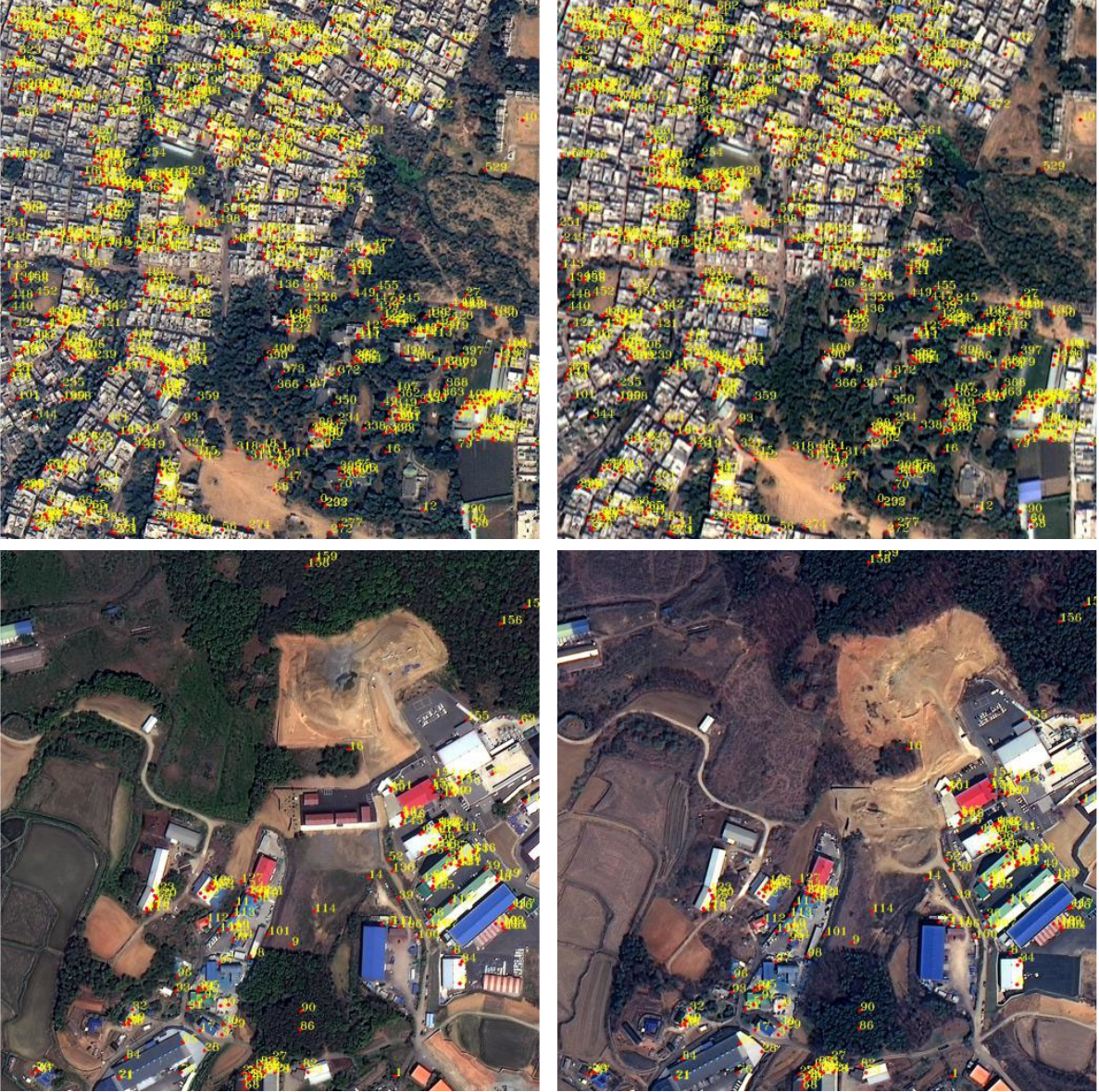}
   \caption{Registered feature points. The images in the left and right columns form bitemporal pairs. The numbered points are the matched feature points in the two images.}
\label{fig:registration}
\end{figure}

{\textbf{Annotation and Quality Control.}}
The bitemporal images were also annotated by the remote sensing image-interpretation
experts. All newly built and demolished building regions in the dataset were annotated at
the pixel level in separate auxiliary graphs, thus making further registration processes
possible. The task was jointly completed by 5 remote-sensing experts from the authors'
research institution, 8 postgraduate students from partner universities and more than 20
well-trained annotators employed by the Madacode AI data service {company} 
 (\url{www.madacode.com}, {Accessed 1 November 2021}). The remote-sensing experts have been undertaking military
 surveillance and damage assessment for natural disasters such as fires for many years. The
 annotation accuracy was required to be higher than 2 pixels. All annotators had rich
 experience in interpreting remote-sensing images and a comprehensive understanding of
 the change-detection task. They followed detailed specifications made by the remote-sensing
 experts for annotating the images to yield consistent annotations. Moreover, each sample in
 our dataset was annotated by one annotator and then double-checked by another to
 maximize the quality of the annotations. The labels were checked by the postgraduate
 students, then batches containing 20 bitemporal images were randomly reexamined by the
 remote-sensing experts. It took approximately 2000 person-days to manually annotate and
 review the entire dataset.

We saw some selected samples from the dataset in Figure~\ref{fig:annotation}. It should be noted that the construction of new buildings and the demolition of old buildings was annotated in separately labeled images. In this way, if the registration accuracy is improved, these labeled images can be simply adjusted without the need for any further annotation.

{\textbf{The S2Looking Dataset.}}
As noted previously, the dataset can be downloaded {from}
\url{https://github.com/S2Looking/}, with our hope being that innovative use of the dataset
in the image-processing and change-detection communities will give rise to new
requirements that can then be addressed through further dataset refinements.


\subsection{The \text{S2Looking} Challenge}\label{sec:challenge}

Buildings are representative manmade structures. During the last few decades, the areas
from which our images were collected have seen significant land-use changes, especially in
terms of rural construction. This presents more difficulties for change detection than
construction in urban areas. Our VHR remote-sensing images provide an opportunity for
researchers to analyze more subtle changes in buildings than simple construction and
destruction. This might include changes from soil/grass/hardcore to building expansion and
building decline. The S2Looking dataset therefore presents current deep learning techniques
with a new and significantly more challenging resource, the use of which will likely lead to
important innovation. To further promote change-detection research, we are currently
organizing a competition based on the S2looking dataset ({see}
\url{https://www.rsaicp.com} Accessed 1 November 2021), which will be addressed to the challenges identified below.


The S2Looking dataset was extracted from side-looking rural-area remote-sensing images,
which makes it an expanded version of the LEVIR-CD+ dataset that will inevitably be harder
to work with. Given S2Looking as training data, the challenge for the change-detection
community is to create models and methods that can extract building growth and decline
polygons quickly and efficiently. Furthermore, these models and methods will need to assign
each polygon in a way that will accurately cover the range of building changes, meaning that
each polygon and building region must be matched pixel by pixel.

Many methods have achieved satisfactory results on the LEVIR-CD dataset (F1-score $>$
0.85)~\cite{2020ASpatial,chen2021efficient}, but the S2Looking dataset presents a whole new
set of challenges that change-detection algorithms will need to be able to address. These are
summarized below:

\begin{description} 
\item[Sparse building updates.] {The} changed building instances in the S2Looking dataset are far sparser than those in the LEVIR-CD+ dataset due to the differences between rural and urban areas. Most rural areas are predominantly covered with farmland and forest, while urban areas are predominantly covered with buildings that are constantly being updated. The average number of change instances in S2Looking is 13.184, while the average number of change instances in LEVIR-CD is 49.188~\cite{2020ASpatial}. This makes it more difficult for networks to extract building features during the training process.

\item[Side-looking images.] The S2Looking dataset concentrates on side-looking
remote-sensing imagery. This makes the change-detection problem different from ones
relating to datasets consisting of Google Earth images. The buildings have been imaged by
satellites from different sides and projected along varying off-nadir angles into 2D images, as
we see in Figure~\ref{fig:side-look}. This means that a change-detection model is going to
have to identify the same building imaged from different directions and detect any updated
parts.

\item[Rural complexity.] Seasonal variations and land-cover changes unrelated to building
updates are more obvious in rural areas than in urban areas. Farmland is typically covered
by different crops or withered vegetation, depending on the season, giving it a different
appearance in different remote-sensing images. A suitable change-detection model needs to
distinguish building changes from irrelevant changes in order to generate fewer
false-positive pixels.

\item[Registration accuracy.] The registration process for the bitemporal remote-sensing
images in S2Looking is not completely accurate due to the side-looking nature of the images
and terrain undulations. Based on the manual screening by experts, we have managed to
bring the registration accuracy to 8 pixels or better, but this necessitates a change-detection
model that can tolerate slightly inaccurate registration.
\end{description}

\subsection{Challenge Restrictions}
To better accommodate operational use cases and maintain fairness, the geographic coordinate information has been removed from the data
used for inference in the challenge.
Thus, no geographic base map database can be applied in the change-detection challenge.
The models for change detection and subsequent image registration are only allowed to extract information from the images themselves.

\section{Results}\label{sec:benchmark}

In order to provide a benchmark against which change-detection algorithms can assess their
performance with regard to meeting the above challenges, we have established a set of
evaluation metrics. To gain a sense of the extent to which S2Looking has moved the
requirements imposed on change-detection algorithms beyond those posed by the current
baseline dataset,  LEVIR-CD+, we undertook a thorough evaluation of benchmark and
state-of-the-art deep-learning methods using both LEVIR-CD+ and S2Looking. This exercise
confirmed the scope of our dataset to establish a new baseline for change detection. We
report below upon the benchmark metrics and the evaluation that was undertaken. We
conclude this section with a close analysis of the evaluation results and the ways in which
existing change-detection algorithms are failing to meet the identified challenges.
\subsection{Benchmark Setup}

{\textbf{Train/Val/Test Split Statistics.}} We evaluated the performance of four classic
(FC-EF, FC-Siam-Conc, FC-Siam-Diff, and DTCDSCN) and five state-of-the-art
(STANet-Base, STANet-BAM, STANet-PAM, CDNet, and BiT) deep-learning methods on
both the LEVIR-CD+ dataset and S2Looking dataset. The specific state-of-the-art methods
were chosen because they had previously performed well for change detection on the
LEVIR-CD+ dataset. As noted in Section~\ref{sec:dataset}, the LEVIR-CD+ dataset contains
985 image patch pairs. We designated 65\% of these pairs as a training set and the remaining
35\% as a test set. This is consistent with our previous work on LEVIR-CD+. The S2Looking
dataset consists of 5000 image patch pairs, which we split into a training set of 70\%, a
validation set of 10\%, and a test set of 20\%. Strictly speaking, a validation set is not
necessary for the training process and it can be merged with the training set. However, the
validation set was able to capture the effect of each iteration during the training process, as
shown in Figure~\ref{fig:results-F1}. After each iteration, the algorithm was tested on the
validation set to assess the level of convergence. In view of the relative difficulty of the
S2Looking dataset, it was felt that it would need a larger training set. The relative
proportions of 70\%, 10\%, and 20\% are also widely used in other deep-learning-based
studies.

{\textbf{Evaluation Metrics.}} In remote sensing change detection, the goal is to infer
changed areas between bitemporal images. To this end, we took three-channel multispectral
image pairs as the input, then output a single-channel prediction map. The Label 1 and Label
2 maps, when combined into one map, form a pixel-precise ground-truth label map. The
performance of a change-detection method is reflected in the differences between the
prediction maps and the ground-truth maps. To evaluate an algorithm's performance,
Precision, Recall, and F1-scores can be used as evaluation metrics:
\begin{align}
  \label{equ:Evaluation} \text{Precision} &=\frac{TP}{TP+FP}, \\
  \text{Recall} &=\frac{TP}{TP+FN}, \\
  \text{F1-score} = &\frac{2}{\text{Precision}^{-1} + \text{Recall}^{-1}}
\end{align}

Here, $TP$, $FP$, and $FN$ correspond to the number of true-positive, false-positive, and
false-negative predicted pixels for class 0 or 1. These are standard metrics used in the
evaluation of change-detection algorithms~\cite{2005Image}. Both Precision and Recall have
a natural interpretation when it comes to change detection. Recall can be viewed as a
measure of the effectiveness of the method in identifying the changed regions. Precision is a
measure of the effectiveness of the method at excluding irrelevant and unchanged structures
from the prediction results. The F1-score provides an overall evaluation of the prediction
results; the higher the value, the  better.

\textbf{Training Process.} For their implementation, we followed the default settings for each of the classic and state-of-the-art methods. Due to memory limitations, we kept the size of the original input images to $1024\times1024$ for the classic methods (FC-EF, FC-Siam-Conc, FC-Siam-Diff, and DTCDSCN) and cropped the images to $256\times256$ for the state-of-the-art methods (STANet-Base, STANet-BAM, STANet-PAM, CDNet, and BiT).
{
It can be seen from Figure~\ref{fig:results-F1} that the performance of the model improves when iteration number N increases. We also observe that the marginal benefit on the model performance is declining with the number of instances increasing. Moreover, there is an upper limit for N when the image does not have more space to superimpose more building instances.
Therefore, depending on the recommended settings for each of the methods, we set
maximum number N = 50 for the classic methods and N = 180 for the state-of-the-art
methods.
}
We trained the detection models on a system with a Tesla P100 GPU accelerator and an RTX 2080 Ti graphics card. All of the methods were implemented with PyTorch.

\begin{figure}[H]
\includegraphics[width=0.98\linewidth]{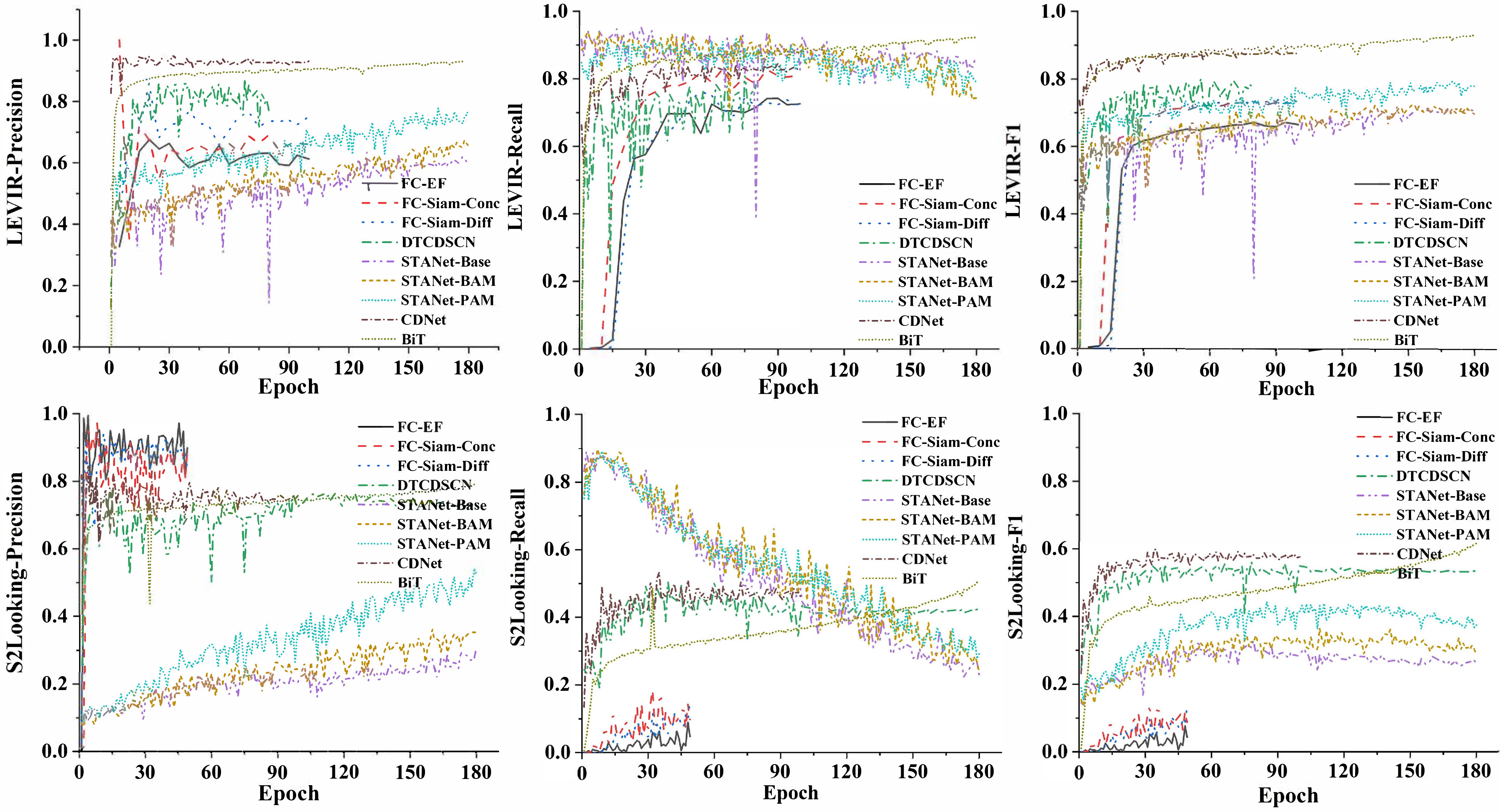}
   \caption{Results of fine-tuning the algorithms on the LEVIR-CD+ and S2Looking datasets.
   The top row presents the Precision, Recall, and F1-score metrics for the LEVIR-CD+ dataset.
   The bottom row presents the metrics for the S2Looking dataset.
   }
\label{fig:results-F1}
\end{figure}

\subsection{Benchmark Results}\label{sec:results}
Figure~\ref{fig:results-F1} shows the results of fine-tuning the various change-detection
algorithms on the LEVIR-CD+ and S2Looking datasets.

Each algorithm took more epochs to converge and obtained lower F1-scores on S2Looking than on LEVIR-CD+. The evaluation metrics and visualizations of the results of the remote sensing change detection for all methods and dataset categories are presented in Table~\ref{tab:experiment} and Figure~\ref{fig:Visualization}, respectively. The F1-scores of the evaluated methods for the S2Looking dataset were at least 25\% lower than the scores for LEVIR-CD+. This confirms that S2Looking presents a far more difficult challenge than LEVIR-CD+.

We also conducted an experiment where only a subset of  the pixels with angles close to on-nadir, i.e., $\pm 15^{\circ}$, were used.
This subset contained 3597 image pairs. The evaluated methods performed much better on
the on-nadir subset of S2Looking, helping us to quantify the effect of using strongly off-nadir
pixels. The evaluation metrics for the on-nadir subset are presented in
Table~\ref{tab:on-nadir experiment}. The F1-scores of the evaluated methods were about
46.4\% greater than their average scores for the overall dataset. This confirms that the
principal difficulty confronting change-detection algorithms when using the S2Looking
dataset arises from the side-looking images.

\begin{specialtable}[H]
\setlength{\tabcolsep}{3.3mm}
\caption{Results for the evaluated methods.}
\label{tab:experiment}
\scalebox{.9}[0.9]{\begin{tabular}{ccccccc}
\toprule
\multirow{2}{*}{\textbf{Method}\vspace{-5pt}} & \multicolumn{3}{c}{\textbf{LEVIR-CD+}} & \multicolumn{3}{c}{\textbf{S2Looking}} \\ \cmidrule{2-7}
 &
  \multicolumn{1}{c}{\textbf{Precision}} &
  \multicolumn{1}{c}{\textbf{Recall}} &
  \multicolumn{1}{c}{\textbf{F1-Score}} &
  \multicolumn{1}{c}{\textbf{Precision}} &
  \multicolumn{1}{c}{\textbf{Recall}} &
  \multicolumn{1}{c}{\textbf{F1-Score}} \\ \midrule
FC-EF\cite{8518015}             & 0.6130   & 0.7261   & 0.6648  & 0.8136  & 0.0895  & 0.0765 \\
FC-Siam-Conc \cite{8518015}     & 0.6624   & 0.8122   & 0.7297  & 0.6827  & 0.1852  & 0.1354 \\
FC-Siam-Diff \cite{8518015}     & 0.7497   & 0.7204   & 0.7348  & 0.8329  & 0.1576  & 0.1319 \\
DTCDSCN \cite{2020Building}           & 0.8036   & 0.7503   & 0.7760  & 0.6858  & 0.4916  & 0.5727 \\
STANet-Base \cite{2020ASpatial}       & 0.6214   & 0.8064   & 0.7019  & 0.2575  & 0.5629  & 0.3534 \\
STANet-BAM \cite{2020ASpatial}        & 0.6455   & 0.8281   & 0.7253  & 0.3119  & 0.5291  & 0.3924 \\
STANet-PAM \cite{2020ASpatial}        & 0.7462   & 0.8454   & 0.7931  & 0.3875  & 0.5649  & 0.4597 \\
CDNet \cite{Chen2021}  & 0.8896 & 0.7345 & 0.8046 &  0.6748  & 0.5493  &  0.6056 \\

BiT \cite{chen2021efficient}  & 0.8274 & 0.8285  & 0.8280 &  0.7264  & 0.5385  &  0.6185 \\
\bottomrule
\end{tabular}}
\end{specialtable}


\vspace{-6pt}
\begin{figure}[H]
\includegraphics[width=0.98\linewidth]{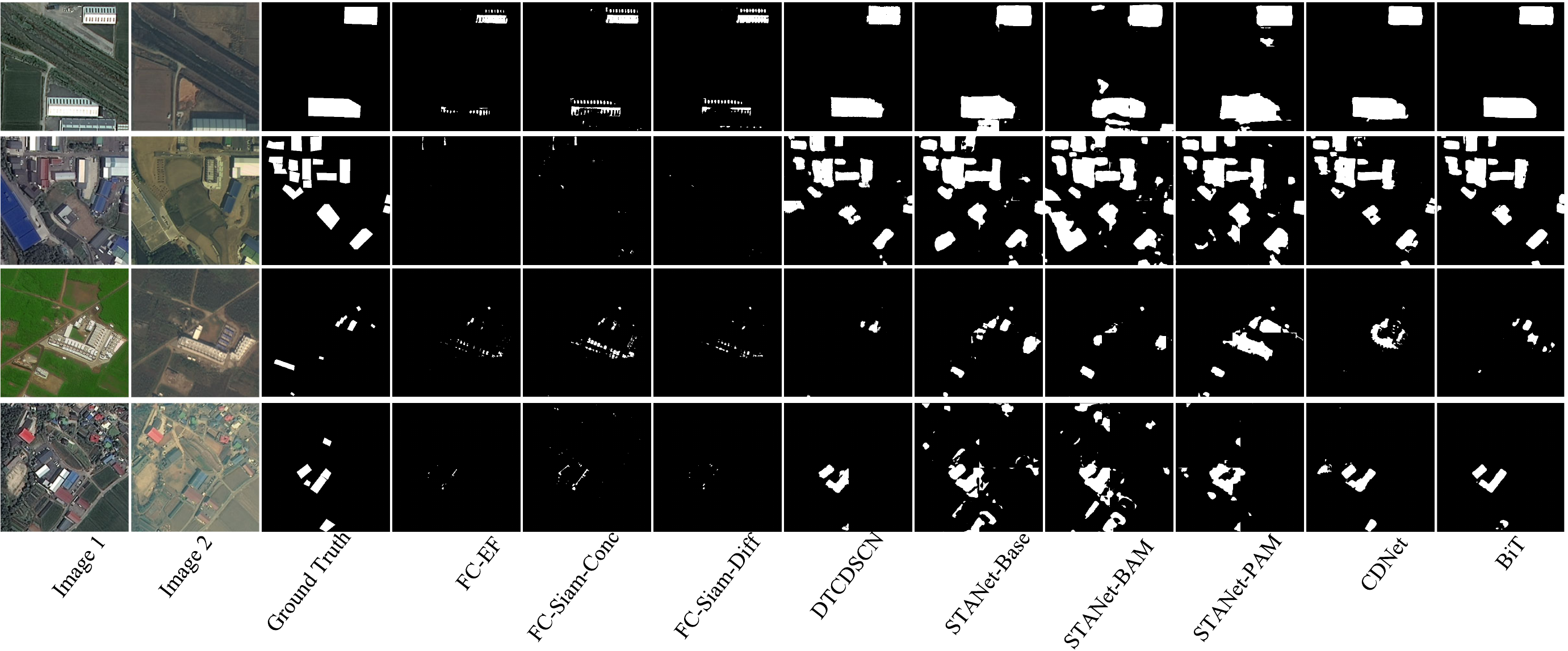}
   \caption{Visualizations of the results of the different methods on the S2Looking dataset.}
\label{fig:Visualization}
\end{figure}

\begin{specialtable}[H]
\setlength{\tabcolsep}{11.4mm}
\caption{Evaluation metrics of on-nadir subset of S2Looking.}
\label{tab:on-nadir experiment}
\scalebox{.9}[0.9]{\begin{tabular}{cccc}
\toprule
             & \textbf{Precision} & \textbf{Recall} & \textbf{F1-Score} \\ \midrule
FC-EF        & 0.7605    & 0.1155 & 0.1825   \\
FC-Siam-Conc & 0.7461    & 0.1663 & 0.2541   \\
FC-Siam-Diff & 0.6609    & 0.115  & 0.1749   \\
DTCDSCN      & 0.7403    & 0.6155 & 0.6436   \\
STANet-Base  & 0.3852    & 0.7344 & 0.4822   \\
STANet-BAM   & 0.4366    & 0.7215 & 0.5206   \\
STANet-PAM   & 0.5103    & 0.7477 & 0.5865   \\
CDNet        & 0.8036    & 0.7375 & 0.7545   \\
BiT          & 0.8512    & 0.733  & 0.7738   \\ \bottomrule
\end{tabular}}%
\end{specialtable}

\section{Discussion}\label{sec:analysis}
A comparison of the results of the evaluation on the benchmark set for the two datasets is
able to reveal the current failings of various algorithms with regard to how they handle the
challenges posed by S2Looking. This, in turn, can provide insights into how they might be
improved. We, therefore, now look at the evaluation results for each \mbox{individual
method.}
%

{\textbf{FC-Net}}~\cite{8518015} consists of three different models, namely, FC-EF, FC-Siam-Conc, and
 FC-Siam-Diff. As shown in Table 3, FC-Siam-Conc and FC-Siam-Diff performed better on
 both datasets than FC-EF. FC-Net performed poorly on S2Looking compared to LEVIR-CD+.
 This is because the structure of FC-Net is too simple to be effectively trained on the complex
 problems presented by the S2Looking dataset. Unlike the DTCDSCN model, which has 512
 channels, the deepest layer of FC-Net has only 128 channels. As a result, the ability of
 FC-Net to capture feature representations is limited. From the predictions produced for the
 S2Looking test set, we can see that the change-detection performance of FC-Net was largely
 premised upon image contrast, which fails to recognize building structures. This results in
 the detection of small objects with strong contrast, such as white cars, darkened windows,
 and the shadows of changed buildings, rather than whole building boundaries. This is
 evident in Figure~\ref{fig:Visualization}.
%

\textbf{DTCDSCN}~\cite{2020Building} is also a Siamese network, but it combines the task
of change detection with semantic detection. DTCDSCN contains a change-detection module
and two semantic segmentation modules. It also has an attention module to improve its
feature-representation capabilities. Compared with FC-Net, DTCDSCN was better able to
identify changed regions and was more robust to side-looking effects and building shadows.
In addition, DTCDSCN was better at detecting small changes in building boundaries
between the bitemporal images. Therefore, DTCDSCN performed much better than FC-Net
on the S2Looking dataset. However, DTCDSCN failed to recognize a number of small
prefabricated houses, as can be seen in the third and fourth rows of
Figure~\ref{fig:Visualization}. Additionally, because of the complexity of the rural scenes,
some large greenhouses, cultivated land, and hardened ground were misrecognized as
changed buildings.

\textbf{STANet}~\cite{2020ASpatial} is a Siamese-based spatiotemporal attention network
designed to explore spatial--temporal relationships. Its design includes a base model
(STANet-Base) that uses a weight-sharing CNN to extract features and measure the distance
between feature maps to detect changed regions. STANet also includes a basic
spatiotemporal attention module (STANet-BAM) and a pyramid spatiotemporal attention
module (STANet-PAM) that can capture multiscale spatiotemporal dependencies. As shown
in Table~\ref{tab:experiment}, STANet-PAM performed better (for its F1-score) than
STANet-BAM and STANet-Base on both datasets. We also found that STANet had a
relatively high Recall but low Precision compared to other methods. This may be because the
batch-balanced contrastive loss that it employs in the training phase gives more weight to the
misclassification of positive samples (change), resulting in the model having a tendency to
make more positive predictions. Note that STANet-PAM performed better than DTCDSCN
on the LEVIR-CD+ dataset but worse than DTCDSCN on the S2Looking dataset. From this,
we conclude that STANet is more vulnerable to side-looking effects and illumination
differences, which are more severe in the S2Looking dataset. Thus, it was more frequently
misrecognizing the sides of buildings as building changes, which influenced the FP value in
Eq.~\ref{equ:Evaluation} and reduced its Precision.

\textbf{CDNet}~\cite{Chen2021} is a simple, yet effective, model for building-change
detection. It uses a feature extractor (a UNet-based deep Siamese fully convolutional
network) to extract image features from a pair of bitemporal patches. It also has a metric
module to calculate bitemporal feature differences and a relatively simple classifier (a
shadow fully convolutional network), which can produce change-probability maps from the
feature difference images. CDNet produced better detection results than the previous
methods on both datasets. This may be because the structure of CDNet, including its
deep-feature extractor, can better handle moderate illumination variances and small
registration errors, enabling it to produce high-resolution, high-level semantic feature maps.
Note in Figure~\ref{fig:Visualization}, for instance, that CDNet was robust in relation to
misregistered hilly regions. However, it failed to predict some small prefabricated houses
and structures that appeared bright in one bitemporal image and dim in another, which were
misrecognized as changed buildings. CDNet has a pixelwise change-discrimination process
that is performed on the two feature maps. This model is not well-equipped to deal with
large side-looking angles.
%

\textbf{BiT}~\cite{chen2021efficient} is an efficient change-detection model that leverages
transformers to model global interactions within the bitemporal images. As with CDNet, the
basic model~\cite{chen2021efficient} has a feature extractor and a prediction head. Unlike
CDNet, however, it has a unique module (a bitemporal image transformer) that can enhance
features. BiT consists of a Siamese semantic tokenizer to generate a compact set of semantic
tokens for each bitemporal input, a transformer encoder to model the context of semantic
concepts into token-based spacetime, and a Siamese transformer decoder to project the
corresponding semantic tokens back into the pixel space and thereby obtain refined feature
maps. Table \ref{tab:experiment} shows that BiT outperformed all the other methods on the
two datasets. Only small prefabricated houses on hills were misrecognized, due to the lower
registration accuracy, as can be seen in the fourth row of Figure~\ref{fig:Visualization}. Most
incorrect pixel predictions produced by BiT were due to side-looking effects associated with
the expansion of building boundaries, which make building boundaries harder to accurately
recognize in remote-sensing images.

Consequently, more sophisticated change-detection models are going to be needed to
efficiently tackle the challenges posed by the S2Looking dataset. It should be noted that the
change-detection methods {here evaluated} were all basically robust to seasonal and
illumination variations in S2Looking, therefore building-change detection using the dataset
is a solvable problem.

\section{Conclusions}\label{sec:conclusion}

This paper has introduced the S2Looking dataset, a novel dataset that makes use of the
camera scroll facility offered by modern optical satellites to assemble a large collection of
bitemporal pairs of side-looking remote-sensing images of rural areas for building-change
detection. The overall goal was to create a new baseline against which change-detection
algorithms can be tested, thereby driving innovation and leading to important advances in
the sophistication of change-detection techniques. A number of classic and state-of-the-art
change-detection methods were evaluated on a benchmark test that was established for the
dataset. To assess the extent to which the S2Looking dataset has added to the challenges
confronting change-detection algorithms, deep-learning-based change-detection methods
were applied to both S2Looking and the expanded LEVIR-CD+ dataset, the most challenging
change-detection dataset previously available. The results show that contemporary
change-detection methods perform much less well on the S2Looking dataset than on the
LEVIR-CD+ dataset (by as much as 25\% lower in terms of their F1-score). Analysis of the
performance of the various methods enabled us to identify potential weaknesses in their
change-detection pipelines that may serve as a source of inspiration for further
developments.

There are several things that could to be actioned to improve upon the S2Looking dataset.
First of all, the level of variation in the captured image pairs makes it difficult for a
registration algorithm to match feature points between them. Improving current registration
techniques would enhance the performance of change-detection models trained on the
dataset. We therefore hope to see S2Looking driving a move towards innovative new
image-registration methods and models. There are also flaws in current change-detection
models, whereby irrelevant structures such as greenhouses and small reservoirs can be
misidentified as building changes. Generally, by bringing this work to the attention of
image-processing researchers with an interest in change detection, we feel that refinement of
both the dataset and existing change-detection methodologies can be driven forward.

\vspace{6pt}

\authorcontributions{ All coauthors made significant contributions to the manuscript.
L.S. conceptualized the problem and the technical framework, developed the collection of
bitemporal data and analyzed the benchmark results, and wrote the paper.
Y.L. conceptualized the problem and the technical framework, ran the annotation program,
selected locations for satellite images, designed the collection of bitemporal data and
analyzed the dataset, developed the collection of bitemporal data, and made the
specifications in the annotation.
H.C. helped conceptualize the problem and the technical framework, helped develop
specifications in annotation and trained annotator from AI data processing company, and
conducted benchmark experiments with CDNet and BiT methods.
H.W. ran benchmark experiments with FC-Net and DTCDSCN and analyzed the data, and
managed training and test datasets for both LEVIR-CD+ and S2Looking.
D.X. ran benchmark experiments with STANet and analyzed data, and conducted the
process of registering the bitemporal images in the S2Looking Data Processing Pipeline.
J.Y. ran benchmark experiments and analyzed data, and completed the process of registering.
R.C. gave advice on the change detection problem,  and reviewed and edited the submitted
version of the article.
S.L. make specifications in annotation and reexamined the quality of the dataset annotations.
B.J. gave trained annotators from the AI data processing company to annotate the dataset,
led the project and gave final approval of the version to be published.
All authors have read and agreed to the published version of the manuscript.}

\funding{{This research received no external funding.}}

\dataavailability{{Publicly available datasets were analyzed in this study. The datasets can be found here: \url{https://github.com/S2Looking/}}.}



\acknowledgments{We would like to thank Ao Zhang and Yue Zhang for collecting the rich satellite images.
We thank Chunping Zhou, Dong Liu, Yuming Zhou from Beijing institution of remote sensing for helping on making specifications in annotation; We also thank Ying Yang and Yanyan Bai from Madacode company for their efforts in the annotation process.}


\conflictsofinterest{The authors declare no conflict of interest.}


\end{paracol}
\reftitle{References}


\end{document}